\documentclass[10pt,twocolumn,letterpaper]{article}

\usepackage{cvpr}
\usepackage{times}
\usepackage{epsfig}
\usepackage{graphicx}
\usepackage{amsmath}
\usepackage{amssymb}
\usepackage{float}
\usepackage{subfigure}
\usepackage{multirow}

\cvprfinalcopy 

\def\cvprPaperID{1636} 

\ifcvprfinal\fi
\begin{document}

\title{Human Action Adverb Recognition: ADHA Dataset and A Three-Stream Hybrid Model} 

\author{Bo Pang, Kaiwen Zha, Cewu Lu \footnote{Corresponding Author}\\
Shanghai Jiao Tong University, China\\
{\tt\small pangbo@sjtu.edu.cn,Kevin\_zha@sjtu.edu.cn,lucewu@cs.sjtu.edu.cn }
}

\maketitle

\begin{abstract}
We introduce the first benchmark for a new problem --- recognizing human action adverbs (HAA): ``Adverbs Describing Human Actions" (ADHA). We demonstrate some key features of ADHA: a semantically complete set of adverbs describing human actions, a set of common, describable human actions, and an exhaustive labeling of simultaneously emerging actions in each video. We commit an in-depth analysis on the implementation of current effective models in action recognition and image captioning on adverb recognition, and the results show that such methods are unsatisfactory. Moreover, we propose a novel three-stream hybrid model to deal the HAA problem, which achieves a better result.
\end{abstract}

\section{Introduction}

Computer vision aims to recognize semantic labels in visual data (e.g., images, video). We find these semantic labels are inside our language system. For example, object detection/recognition \cite{ren2015faster, redmon2016yolo9000} can be considered as exploring ``noun'' in visual data. To understand ``verb'', action recognition \cite{donahue2015long, wu2015modeling, srivastava2015unsupervised,yue2015beyond, karpathy2014large,ji20133d,karpathy2014large} has been extensively studied. Moreover, the ``Adjective'' labels (e.g., cool, dark, beautiful) are explored by attribute learning \cite{petrosino2010toward}. Until now, most of computer vision researchers have ignored an important kind of words --- ``\textbf{Adverb}'', which can express the attitude and mood of the subject, and the attributes of the action as well. From the viewpoint of language research \cite{white1991adverb},  these concepts convey more sensitive semantics compared with actions and nouns.

If we can teach an AI to understand adverbs of an action, it implies that the AI can understand the attitude and mood of the action player, which is necessary for interactive robots. We also believe this is the preliminary work to make AI understand the purpose and the intent behind actions.

We are the first one to explore the topic of human action adverbs (HAAs) recognition. Unlike action recognition, HAAs describe the conceptions with very sensitive visual patterns that are difficult to recognize. For example, is the drinking people happy or sad; Is the hand shaking an expression of excitement or politeness? Extensive experiments show that understanding adverb is very challenging to current state-of-the-art deep learning architecture. Note that in the topic of image captioning \cite{xu2015show}, adverb may be included in the language material. However, on one hand, they don't take adverb as a target. On the other hand, we believe adverb recognition will be one of the most important tools to further advance the topic of image captioning.

\begin{figure*}
	\begin{center}
		\includegraphics[width=\textwidth]{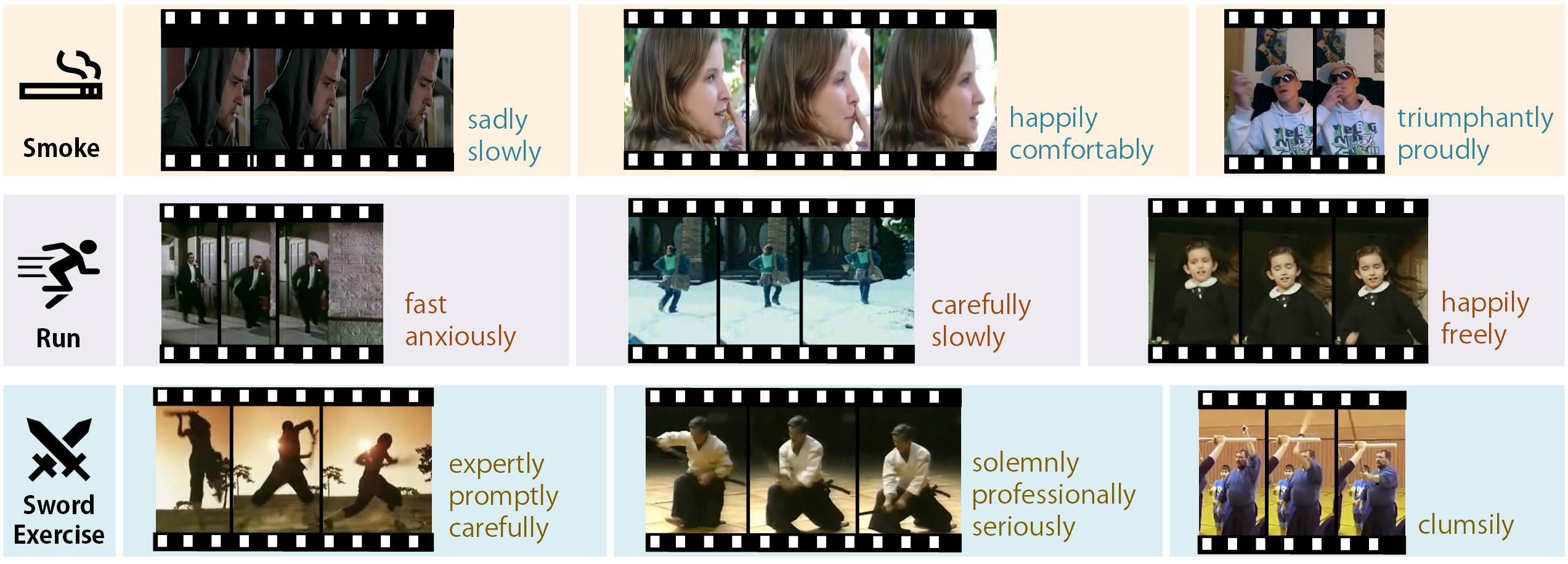}
	\end{center}
	\caption{Example frames and annotations in ADHA.}
	\label{fig:pinjie}
\end{figure*}

By now, we have made tremendous progress in action recognition. Many large-scale datasets like UCF101~\cite{soomro2012ucf101}, YouTube-8M~\cite{abu2016youtube} are built and some excellent models like two-stream network~\cite{simonyan2014two} \etal have been proposed. Despite significant advances in action recognition, HAAs, as an important attribute is a missing study. There is no appropriate dataset for HAA recognition.

Similar to other beginnings of new topics, we built a video dataset with adverbs labeled for actions. There are 12000 videos in total covering 350 action-adverb pairs (e.g.,  ``smoking sadly") over 32 actions described by 51 adverbs. There are three kinds of HAAs that can respectively describe the subject's mood, attitude and the attributes of the actions. We commit a social experiment to make sure that the HAAs space can describe almost all the meanings a human want to express after seeing a short action video.

We highlight four features of the dataset. First, an average of 11 distinct HAAs per action (no dull action which can only be labeled by few adverbs). Second, our adverb categories are based on semantics rather than words. It means that we do not take  ``smoke sadly" and ``smoke sorrowfully" as different categories. Third, the dataset is multi-labeled. An action can be labeled with multiple HAAs describing mood, attitude and action attributes simultaneously. Fourth, each video is labeled by three annotators, with different backgrounds to reduce bias. Accordingly, a novel evaluation metric is designed. Fig.~\ref{fig:pinjie} shows example video frames and annotations in ADHA.

We prove that this dataset is able to act as a benchmark of the HAA recognition problem. We commit experiments to answer the following questions: First, how well can the current action recognition and image captioning approaches deal with the HAA recognition problem? Second, can pose help to understand HAA?  Third, can we use expression knowledge as extra information to solve the HAA recognition problem?

The contributions of this work are that: 1) we build a large-scale video dataset labeled with actions, HAAs, and human instance boundingboxes. 2) We benchmark several current action recognition, pose estimation, and image captioning models on ADHA. 3) We propose a hybrid model using pose, expression, optical flow and RGB information and achieve a relative better performance.

\subsection{Related Dataset}

 To our knowledge, we are the first to study HAA recognition. Building high-quality benchmark dataset is the first step to explore this topic. Therefore, in this section, we focus on investigating some related datasets.
\paragraph{Action Dataset}
Action recognition in video has made great progress due to many excellent datasets, from small simple datasets like KTH~\cite{laptev2005space} to large-scale, real-world datasets such as Youtube-8M~\cite{abu2016youtube}, UCF-101~\cite{soomro2012ucf101}, and Sport-1M~\cite{karpathy2014large}. However, action is not only contained in video. In other words, people as well as AI can tell the action by only one image. Sometimes, recognition of an action in videos stems from the recognition of a related object in the scenario. For example, a model might recognize the action of swimming by recognizing the swimming pool. Actually, the model does not understand, or rather, pay attention to the action itself. Instead, it solves an object detection problem. However, if a model can recognize adverbs very well, we can tell that the model has the ability to understand the action, actions' attributes and the player's attitude. This is why HAA recognition is important and difficult.

\paragraph{Video Captioning Dataset}
Video captioning is a hot issue and there are many datasets designed for it, for example MSR-VTT~\cite{xu2016msr} which is built from the queries on a commercial video search engine and covers 10k clips, YouTube2Text~\cite{guadarrama2013youtube2text} with 2,000 video snippets and 120K sentences, ActivityNet Captions~\cite{Krishna_2017_ICCV}, a large-scale benchmark for dense-captioning events which contains 20k videos amounting to 849 video hours with 100k total descriptions. Although video captioning needs adverbs, it needs to concern many other things like fixed phrases and idioms, which is biased against our goal. And in our HAA set, we have removed the synonyms, because we do not want to discriminate them. While in video captioning, they need to be taken into account. Therefore, although there are many datasets built for video captioning, they are not suitable for HAA recognition.
\paragraph{Human Expression Dataset}
Face expressions can convey the mood and attitude of human like HAAs. And many datasets are built for it like Emotiw2016~\cite{dhall2016emotiw} which contains an audio-video based emotion and a group-based emotion recognition sub-challenges, MMI~\cite{valstar2010induced} --- a resource for building and evaluating facial expression recognition algorithms, HUMAINE~\cite{douglas2007humaine} which provides naturalistic clips recording pervasive emotion and suitable labeling techniques. But there are also some shortcomings. The expression recognition mainly deals with faces in images or videos, but HAA recognition needs to recognize subject's mood and attitude from his action not merely face expression. Moreover, adverb recognition can tell the attributes of the action like how fast, heavily, which is beyond the expression recognition. Therefore, we need to build a HAA recognition dataset to solve this much more difficult and comprehensive problem.

\section{Constructing ADHA}
\begin{figure}
	\begin{center}
		\includegraphics[width=3.2in, height=2.5in]{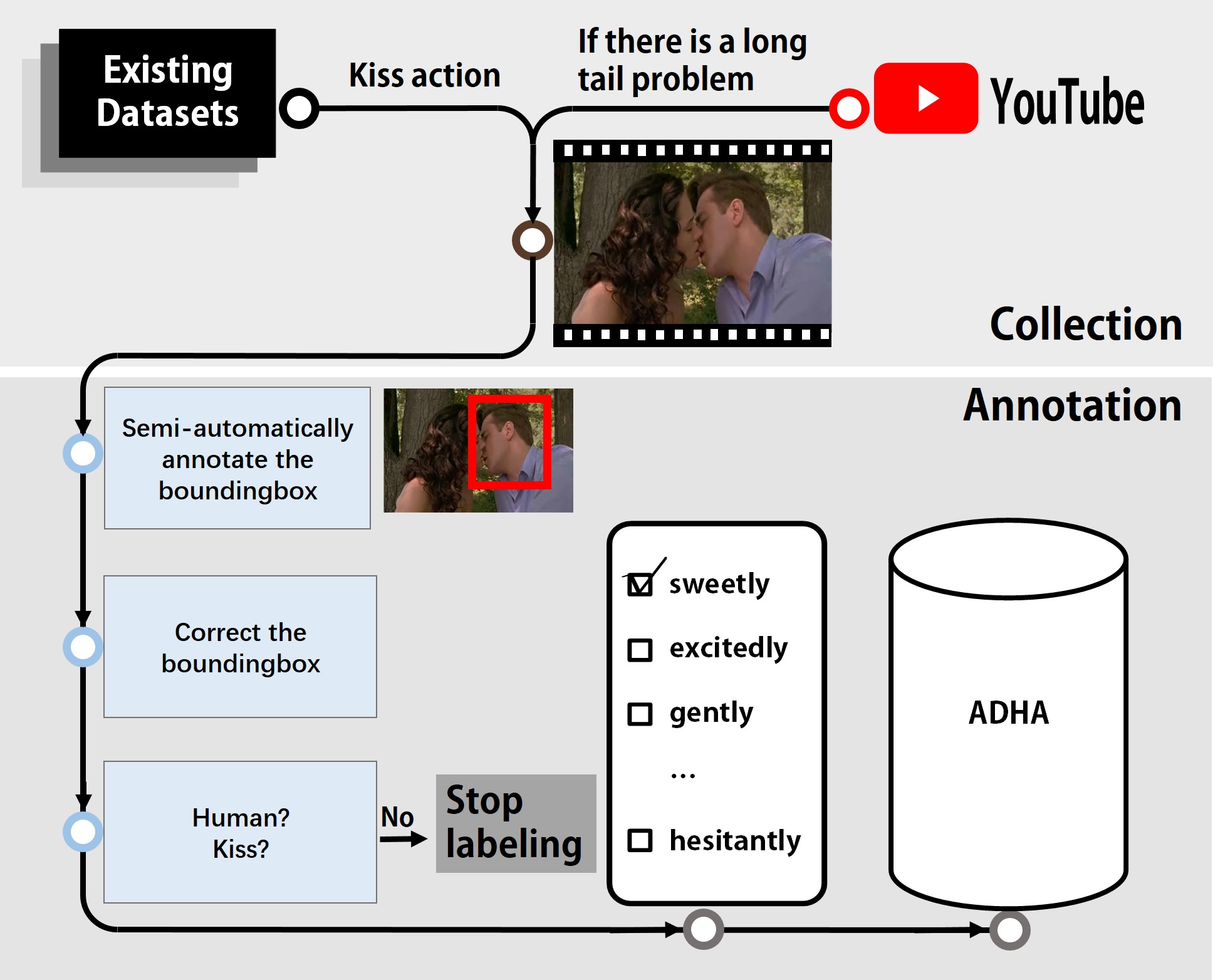}
	\end{center}
	\caption{Pipeline for collecting and annotating the videos. We take ``kiss'' action as an example.}
	\label{fig:pipeline}
\end{figure}

Our dataset is constructed for recognizing adverbs describing human actions (HAA). First, we introduce how to make the action and adverb list. Then, videos are collected based on the given lists. The pipeline of the annotation is shown in Fig.~\ref{fig:pipeline}. Finally, we present the annotation details on the collected videos.
\subsection{Selecting Action and Adverb Categories}
Adverbs are used to describe the actions, so we fix the action categories list first. Then make the adverb list according to the given selected action categories.

\paragraph{Action Collection} Previous action dataset~\cite{abu2016youtube,laptev2005space,soomro2012ucf101, karpathy2014large} constructors seek to build sets that cover most of action categories, those frequently happen in daily life. So, in this paper, we adopt the union set of action categories from these datasets as our action candidates shortlist. As a research dataset, we expect the selected actions to require adverb descriptions. For example, in common cases some sport actions such as fencing, gymnastics, swimming are without attitude and mood. So those adverb-needless actions are not our target. To further refine the shortlist, we turn to language prior knowledge. We extract about 0.2 million video descriptions as our language materials. Then we rank action categories by the percentage of being described by at least one adverb. According to this score, we finally choose top 32 actions in the shortlist:\{brush hair, chew, clap, climb stairs, dive, draw sword, drink, eat, fall floor, hit, hug, kick, kiss, pick, pour, pullup, punch, push, run, shake hands, shoot bow, shoot gun, sit, smoke, stand, swing baseball, sword, sword exercise, talk, throw, walk, wave\}.

\paragraph{Adverb Collection} Given action categories, we build the list of the adverbs. We consult the word frequency from the Corpus of Contemporary American English (COCA) which is an authoritative corpus of American English~\cite{davies2008corpus}. From thousands of adverbs, we choose 113 adverbs which are able to describe actions and possess the highest word frequencies. After removing the synonyms, there are 51 adverbs left. In order to make sure that these 51 adverbs cover all the meanings a person wants to express after seeing a short action video, we conduct a user study. We invite 50 students with different majors from the college and give each of them 50 videos and the adverb set. They need to watch each video first and then check whether the adverb set can cover what they want to express about the video. From the result, we find that for male, the adverb set can cover $98.8\%$ requirements and for female it is $97.4 \%$.

\paragraph{Adverb-action Pair Collection} With 32 actions and 51 adverbs, we then group adverb-action pairs. In this process we take a N-gram data\footnote {https://www.ngrams.info/} as the reference. This N-gram data is better than Google N-Gram for us, since it has a large corpus (COCA) and takes the low frequency (even only appear 1 or 3 times) n-grams into account. This n-gram data provides us the frequencies of action-adverb pairs. Then we obtain the candidate action-adverb pairs set on the basis of the frequencies. For every action there are around 11 appropriate adverbs. In total we have 350 adverb-action pairs. We will list the adverb and adverb-action pairs in supplementary file.

\subsection{Video Collection and Annotation}
\paragraph{Video Collection} We collect video clips from both YouTube and existing action datasets. First, videos from several existing datasets { HMDB51~\cite{kuehne2011hmdb}, HOHA~\cite{laptev2008learning}, UCF-101~\cite{soomro2012ucf101}} are used. Then, we add videos from YouTube in order to deal with the long tail problem on adverbs. The details of video collection will be illustrated in section \ref{sec:stat}. Each human instance in the video preforms one action only.

\begin{figure*}[t]
	\begin{center}
		\includegraphics[width=1\textwidth, height=3in]{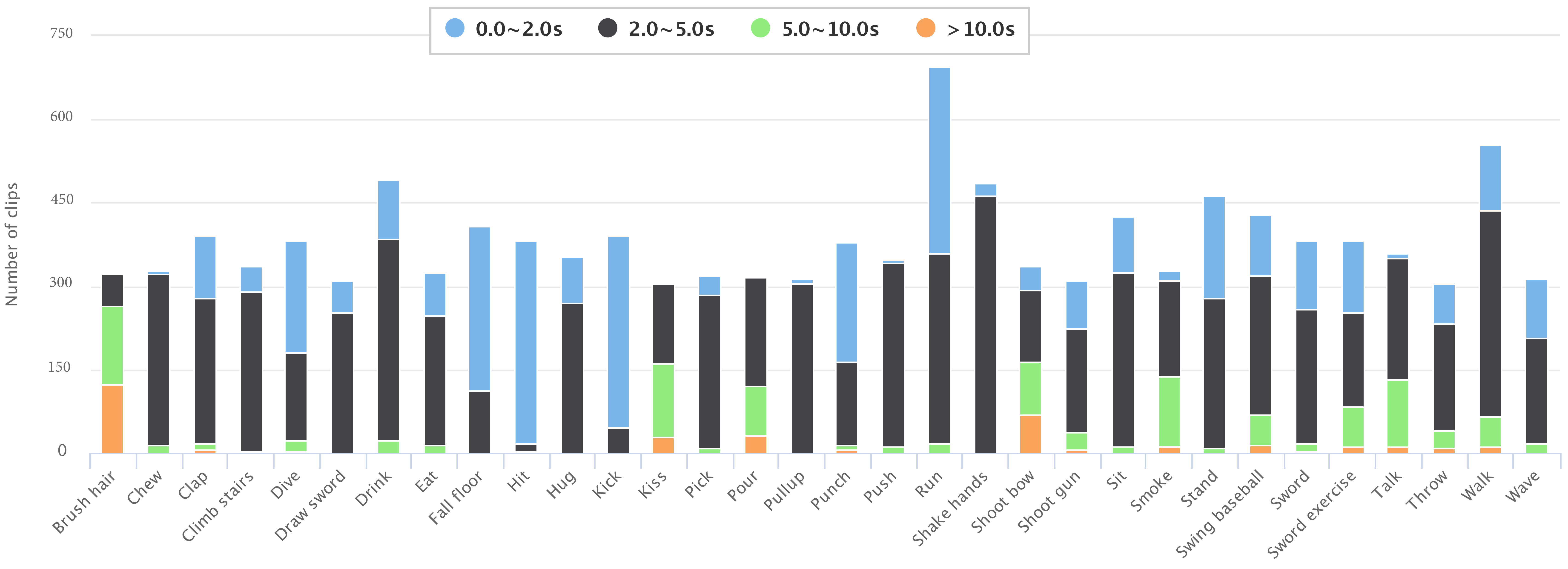}
	\end{center}
	\caption{Number of videos per action. The distribution of video durations is illustrated by the colors.}
	\label{fig:LA1}
\end{figure*}

\paragraph{Human Instance Annotation} Our annotation is in human instance level, because we should know who performs the action. What's more, some actions like ``kiss'' and ``hug'' have more than one player, so that we need to annotate them respectively. We propose a semi-automatic annotation framework to effectively localize human instance.
We label the human boundingbox at the first frame and use object tracking model MDNet~\cite{nam2015mdnet} which is the winner of the VOT-2015 challenge~\cite{Kristan_2015_ICCV_Workshops} to search corresponding human instance in the following frames. To improve the robustness, we implement human detection (using Faster-RCNN~\cite{girshick2015fast}) to revise tracking bounding box. In detail, we pick up a human detection box that is the closest to the tracking result box based on IOU overlap criterion, then average them as revised result. Annotators observe the automatic video annotation on-line. If the automatic annotation is inaccurate, the annotator need to stop the video and manually correct the bounding boxes. In this way, we only need to annotate some key frames, instead of all frames.

\begin{figure*}[t]
	\begin{center}
		\includegraphics[width=1\textwidth, height=3in]{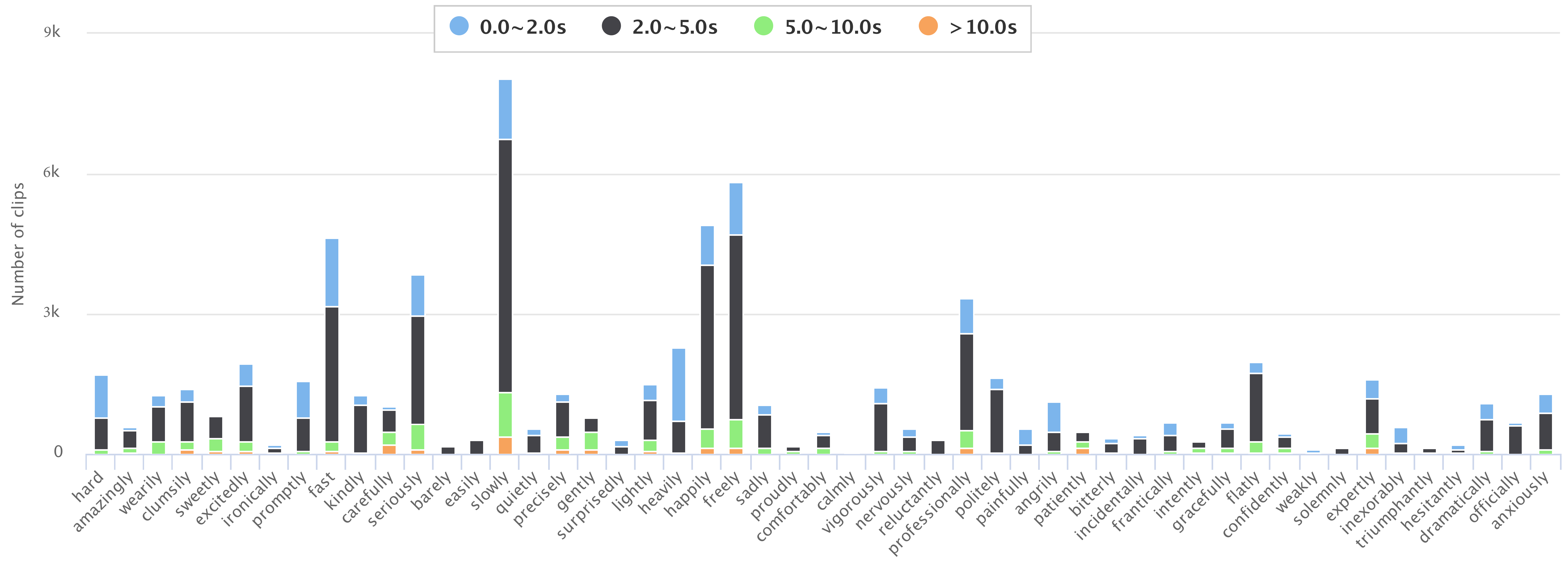}
	\end{center}
	\caption{Number of videos per adverb. The distribution of video durations is illustrated by the colors.}
	\label{fig:LA2}
\end{figure*}

\paragraph{Annotator} We invite 100 annotators with different ages, genders and nationalities. Since our adverbs are presented in English, all the annotators are either native English speakers or excellent English speakers.

\paragraph{Annotation Interface} Our interface is friendly. We play video with labeled human instance boxes. Then, the system gives out an adverb list to choose from. Annotators can select one or multiple adverbs to describe the observed action video, and they can replay it many times until they are confident enough with their choices. We don't set any time limitation for them.

\paragraph{Work Assign} Adverbs are used to describe mood, attitude like concepts which are more subjective than object and action category labeling. Annotating by only one person may not cover all the feelings of people. Therefore, each video is assigned to three different annotators. We make sure those three annotators should have diverse backgrounds (e.g., nationality, age, gender, education background). We find three annotations for each action instance is enough to cover most of feelings. A study experiment is conducted: After annotation is done, we randomly sample 1000 action instances and re-label them by 30 new annotators outside those 100 annotators. We find that $96.3\%$ cases have at least one original annotation which is exactly the same with the new one.

\subsection{Dataset Statistics and Discussion}\label{sec:stat}

In total, there are 32 actions, 51 adverbs, 350 action-adverb pairs, and 12000 videos of ``.avi" in ADHA. A video may be annotated with more than one adverb and the average number of adverbs per video is 1.81. Some statistics about the videos are shown in Tab.~\ref{tab:sav}. We can tell that the dataset contains only short videos, which reduces the difficulty. For every action and adverb, we count the amount of the videos about them and show the distributions in Fig.~\ref{fig:LA1} and Fig.~\ref{fig:LA2}.  To deal with the long tail problem, we recollect many videos. For example, before recollection, ``hug sadly" only has 9 samples in 483 videos. While after recollection, it has 30 samples in 510 videos. To get the diversity among the 3 annotations, we use a 51-dimension vector to represent the annotation, then calculate the average of the Manhattan distances between every two of the 3 annotators. Using $a_{v,i}$ to represent the $i$-th annotation of the video $v$, then the diversity $d$ can be written as:
\begin{equation}
d=avg_v(\sum_{i<j}{|a_{v,i}-a_{v,j}|])}.
\end{equation}
The diversity of ADHA annotation is 1.376.


\begin{table}[H]
	\begin{center}
		\begin{tabular}{|c|c|}
			\hline
			\#Actions & 32\\
			\#Adverbs & 51\\
			\#Clips & 12000\\
			Mean Clip Length & 3.25s\\
			Total Duration & 43071s\\
			\hline
		\end{tabular}
	\end{center}
	\caption{Statistics about the videos}
	\label{tab:sav}
\end{table}

\section{Benchmark System}
In this section, we evaluate several representative approaches designed for action recognition and image captioning using our ADHA dataset and analyze the influence of the expression knowledge on HAA recognition. First, we will briefly introduce the prior approaches.

\subsection{Related work}
\textbf{Action Recognition:} We have made great progress in action recognition and many excellent models have been proposed. Prior works have explored various strategies. Some works extract features of video frames and then fuse them. RNN is widely used to fuse~\cite{donahue2015long, wu2015modeling, srivastava2015unsupervised} and many pooling methods have been developed~\cite{yue2015beyond, karpathy2014large}. CNN gains great success in image processing so that many CNN models in video field appear like 3D-CNN~\cite{ji20133d} and time dimension convolution~\cite{karpathy2014large}. Whereas other jobs take other methods to deal with the temporal information like optical flow~\cite{simonyan2014two}, trajectories~\cite{wang2011action}, and human pose estimation~\cite{maji2011action}. All of these methods do good job in action recognition on video. Can they deal with the HAA recognition problem as well?

\textbf{Face Expression:} Adverbs convey mood and attitude, which can also be inferred by face expression. Nowadays there are many outstanding expression recognition models like \cite{fan2016video} which uses a C3D, CNN, RNN hybrid networks, and \cite{ng2015deep} which uses transfer learning to deal with small datasets.

\textbf{Pose estimation:} For pose estimation, we usually output the heat map of the key points or the skeleton. In ~\cite{newell2016stacked} a stacked hourglass network is used. And AlphaPose~\cite{fang2017rmpe} consists of three components: Symmetric Spatial Transformer Network (SSTN), Parametric Pose NonMaximum-Suppression (NMS), and Pose-Guided Proposals Generator (PGPG).

\textbf{Image Captioning:} Image captioning also needs adverbs. In~\cite{pan2004gcap} a graph based method is used. In~\cite{xu2015show} the authors add visual attention to the model.

\subsection{Experiment Setup}
\paragraph{Metric} We choose mean average precision (mAP) and Hit@k as the evaluation metrics because a video can be labeled with more than one adverb and the adverbs are not exclusive between each other. On the other hand, mAP and Hit@k are  widely used metrics for action recognition and have been used in many benchmarks~\cite{soomro2012ucf101, abu2016youtube,karpathy2014large}, which will give people a better intuitive feeling on the metric values.

mAP: Given a video, a model will give a classification score for every adverb. Then we compute average precision (AP) for every adverb using the sorted classification scores. The average value of all the adverb APs is mAP.

Hit@k: This measures the fraction of test samples that contain at least one of the ground truth labels in the top $k$ predictions. If $rank_{v,e}$ is the rank of entity $e$ on video $v$ (rank 1 gives to the best scoring entity), and $G_v$ is the set of ground-truth set for $v$, then the $Hit@k$ can be written as:
\begin{equation}
1/|V|\sum_{v\in{V}}{\lor_{e\in{G_v}}\amalg(rank_{v,e}\leq k)},
\end{equation}
where $\lor$ is logical OR and $\amalg$ is indicator function.

\paragraph{Evaluation}


We need to define the positive and negative samples first. We treat every person as one sample instead of one video, since there may be more than one player in one video. If the person in the boundingbox is doing the specific action and the action matches the features of one candidate adverb, then, this is a positive sample. If not, then this is a negative sample for that adverb. But what about the condition where the person in the boundingbox is not doing the specific action? Do we just delete it from the sample set or treat it as a negative sample? We treat it as a negative sample. Because models do not have the prior knowledge that whether the person is doing the specific action, we cannot delete it. We realize that there is a possibility that the person in boundingbox who is not doing that specific action is doing another action which is in the action set. we have conducted an experiment and found that the possibility is almost 0. We randomly choose some boundingboxes belonging to the third group to check whether there is such a risk, and have not found such an example. It is enough to prove that the rate is small enough to ignore.

We use 80\% of the dataset as training set and the remaining as test set. When splitting, we put the videos in the same scene only in one set. Although we deal with the long tail problem, the problem still exists, which is a challenge for HAA recognition approaches.

We use two tasks to evaluate the adverb recognition problem: one is given a video then recognizing the action and adverbs (task 1); the other is given a video and its action categories then recognizing the adverbs (task 2).

Since we have 3 annotators for each case, we use the one which has the closest distance to the predicted value as the ground truth to calculate the evaluation metrics. We use this setting because we believe if a model can recognize HAA like at least one annotator, it is a good model.

\subsection{Attention mechanism}
\begin{figure}
	\begin{center}
		\includegraphics[width=3.2in, height=1.8in]{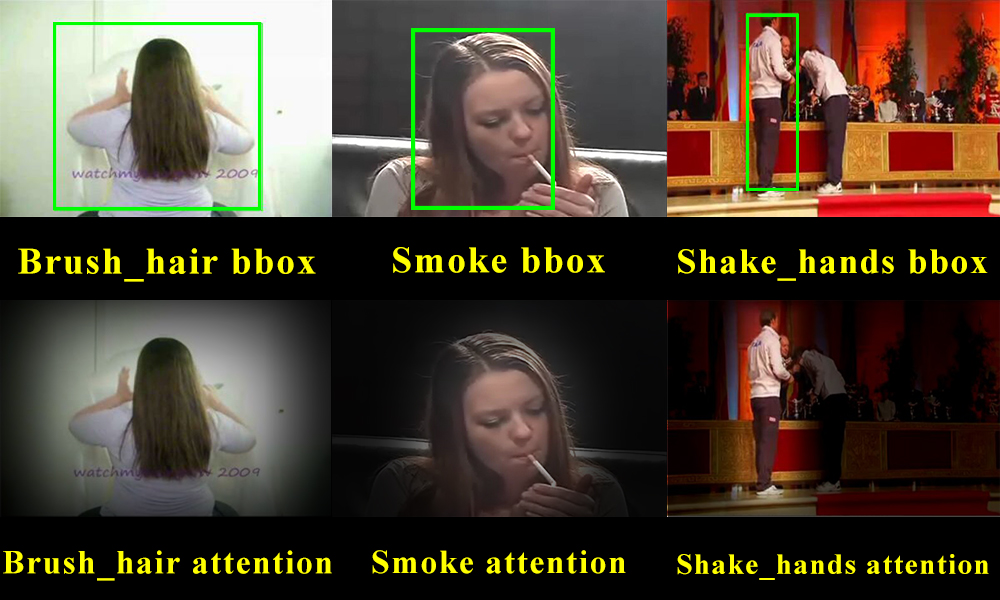}
	\end{center}
	\caption{Examples of attention. Top: Row images with boundingboxes. Bottom: Corresponding images after attention process.}
	\label{fig:attention}
\end{figure}

In this dataset, we treat every person instead of a video as a sample. If we consider human boundingbox regions only, the background context which is useful for HHA recognition may be missed. Therefore, we need to consider the whole image and also tell the model which person in the video to look at. In these experiments, a much more effective attention mechanism is used. We lower the brightness beyond the boundingbox instead of increasing the boundingbox's brightness to reduce the loss of the color saturation in attention area. To avoid fetching in extra edges, we commit smoothing process and some examples are shown in Fig.~\ref{fig:attention}. Smoothing can also deal with the deviation of the MDNet when generating the boundingboxes (like Smoke bbox in Fig.~\ref{fig:attention}). If $ \sigma$ is the value to lower the brightness, $B$ is the attention area, $c$ is the center of $B$ and $a$ is the raw value of the point $p$, then the decayed value of $p$ can be written as: $\max (0, a-|p-c|\amalg(p\notin B)\times\sigma)$. In this way, the target person is indicated from the background context.

\subsection{The Three-Stream hybrid model}
\begin{figure}
	\begin{center}
		\includegraphics[width=3.2in, height=1.8in]{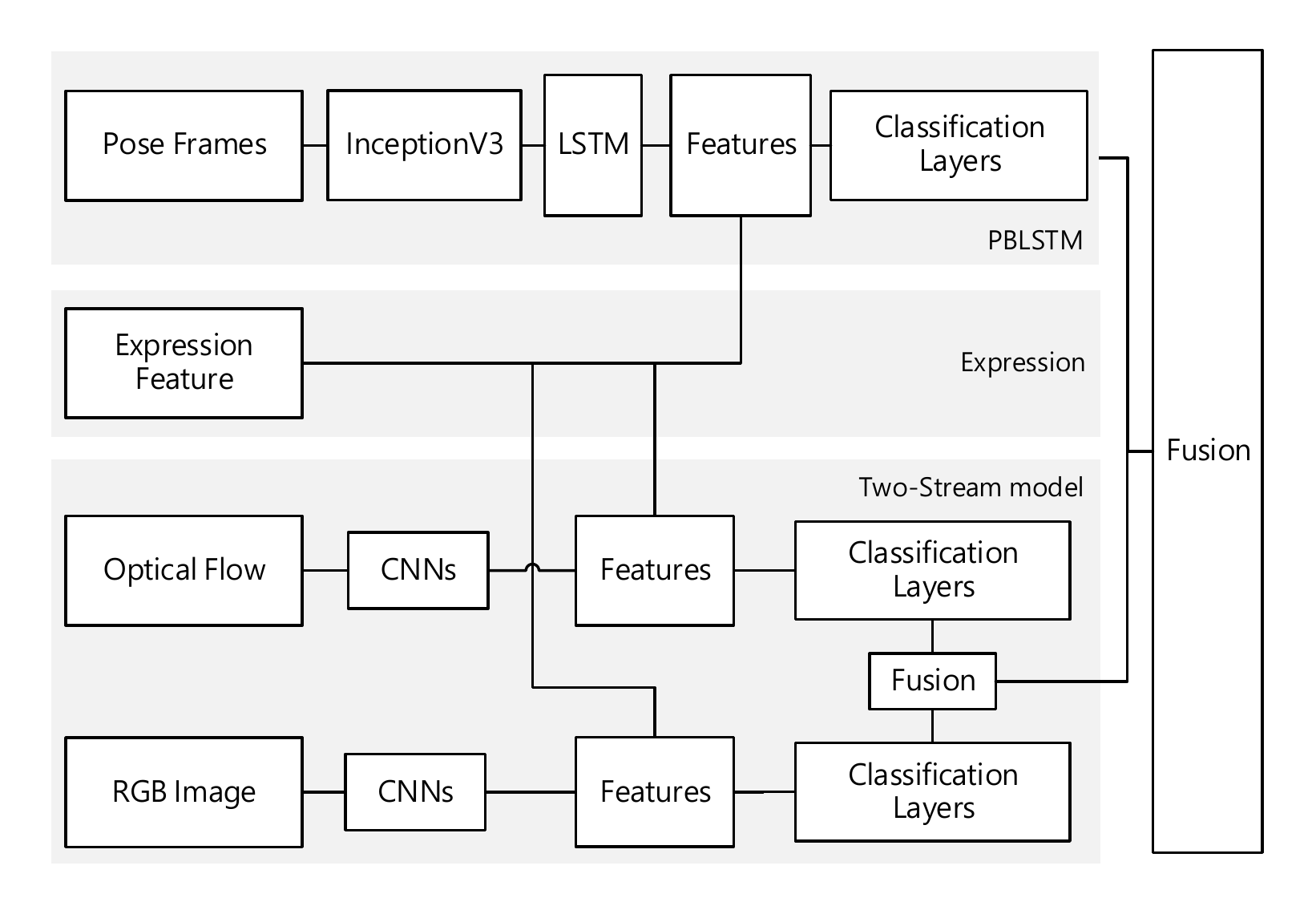}
	\end{center}
	\caption{The framework of the Three-Stream Hybrid model. The PBLSTM stream uses pose information; The expression stream uses expression information; And the Two-Stream stream uses optical flow and RGB.}
	\label{fig:HM}
\end{figure}

We propose a hybrid model using RGB, optical flow, pose and expression information and benchmark it on our ADHA dataset. The framework of the model is shown in Fig.~\ref{fig:HM}.

\subsubsection{Two-Stream Sub-Model}
Two-stream model is really a successful model for action recognition. The two streams are spatial stream and temporal (motion) stream. The former uses frames' RGB information while the latter uses the optical flow information which can show the shifting of every pixel in the video. With these two kinds of information, the model can tell what is in the video and how it moves. We refer to~\cite{simonyan2014two} to implement the two-stream model. Instead of multi-task learning used in~\cite{simonyan2014two} to train the temporal stream, we use cross modality pre-training method proposed in~\cite{wang2016temporal} to do weights shape transform so that we can use the ImageNet pre-trained weights in the temporal stream.

We use OpenCV to extract the dense optical flow where the Gunnar Farneback algorithm~\cite{farneback2003two} is used. In this algorithm, the first step is to approximate each neighborhood of both frames by quadratic polynomials, which can be done efficiently using the polynomial expansion transform. Then from observing how an exact polynomial transforms under translation a method to estimate displacement fields from the polynomial expansion coefficients is derived~\cite{farneback2003two}. After getting the optical flow, we sample 10 frames uniformly from the temporal space as the input.

The spatial and motion stream CNNs are pre-trained ResNet101~\cite{he2016deep} using ImageNet and fine-tuned on ADHA dataset. Then we put the feature maps provided by these CNNs into different classification layers to do different tasks.

\subsubsection{Pose Based LSTM Sub-Model (PBLSTM)}
\begin{figure}
	\begin{center}
		\includegraphics[width=3.2in, height=1.93in]{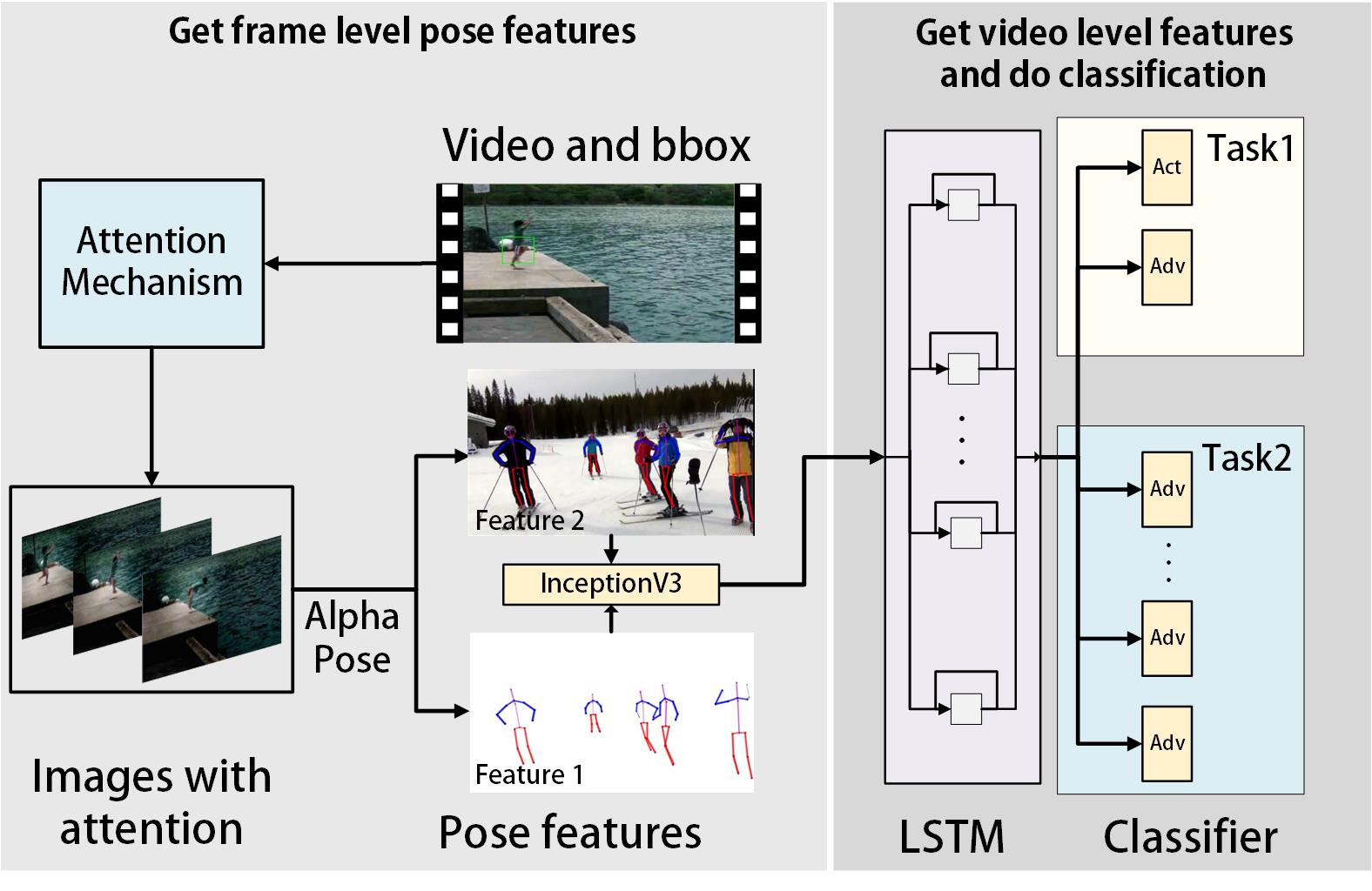}
	\end{center}
	\caption{PBLSTM Pipeline. Get the attention pose features first. Then use LSTM (3 layers with 2048 units) to get the fused video level features. For task 1, action and adverb recognition have their own classifiers (``Act'' and ``Adv'' in the figure). For task 2, every action has its own adverb recognition classifiers. }
	\label{fig:LSTMpipe}
\end{figure}

LSTM models have achieved great results in action recognition problem. We take an approach similar to~\cite{donahue2015long} to utilize LSTM for adverb recognition. However, unlike that work, we do not use the row video frames as the inputs of the CNN. For adverb recognition, pose is a kind of valuable feature. Although we can gain pose information from the raw frames, they are high dimensional which are not easy to use. So we adopt a state-of-art model ~\cite{fang2017rmpe} to extract pose rending frames and the pure pose frames as the input to reduce the difficulty.

AlphaPose is a system for multi-person 2D pose estimation.The “Symmetric STN + SPPE” module receives the human bounding boxes obtained by the human detector as input. For each detected bounding box, the corresponding human pose will be predicted by SPPE. And redundent human poses are eliminated by parametric Pose NMS to obtain the final human poses.

We use inceptionV3~\cite{szegedy2016rethinking} pre-trained on ImageNet~\cite{deng2009imagenet} to extract the feature maps of the poses and we get temporal irrelevant features of each frame. Then put these features into LSTM model to obtain global features and do the recognition. The whole pipeline of the PBLSTM is shown in Fig.~\ref{fig:LSTMpipe}

We set the number of stacked LSTM layers and the number of the hidden units in each layer as the hyper-parameters. And the experiments show that 3 layers with 2048 units in them achieve the best performance.

Because in the dataset all the videos are short videos with length around 5s (3fps), we set the maximum number of frames to be 30 which means that the LSTM model is unrolled for 30 iterations. Although a larger unroll number leads to a better performance, it has much lower efficiency.

For task 1, we need to recognize actions and adverbs simultaneously. We set the classification layers into two parts, one for action recognition, the other for adverb recognition. They share the same LSTM layers's weights. For task 2, each action has their own classification layers to do the adverb recognition task and they also share the LSTM layers. The input of the LSTM model is the feature maps of the pose image, and we use two kinds of pose images: one (Feature 1) is pure pose images which only show the human skeletons without background, the other (Feature 2) is the skeletons rendered with the RGB images. There is a problem that whether the RGB information will provide more useful information or more noises for the HAA recognition.
\subsubsection{Using Expression Knowledge}
As discussed earlier, expression recognition has some intersection with adverb recognition. Although they are different problems with different emphasis, we can use the expression as an extra information to improve the performances of the models. We adopt the model~\cite{fan2016video} which is the winner of EmotiW2016~\cite{dhall2016emotiw}. It uses a C3D, CNN, RNN hybrid networks.

After getting the expression labels of each video, we use it as another feature and combine them with the CNN features to get the final adverb recognition results. We expect that with the expression information, the classifier will do a better job on adverbs about emotion.

\subsubsection{Fusion the Sub Models}
We will show the result of the sub-models in next part. For each sub model we choose the best settings and add the expression features in them. We use average polling method to finally fuse them to get the final results of our hybrid model.


\subsection{Result and Discussion}
First, let's analysis the PBLSTM model's results which are shown in Tab.~\ref{tab:res_PBLSTM}. In the table, ``T1-F1" means task 1 using feature 1 mentioned above. And ``-e" means using expression knowledge.

We can see that for task 1, using the two kinds of features achieves almost the same results and feature 2 is a little better. While for task 2, feature 1 is much better on $Hit@1$ and $Hit@5$. This reveals that RGB information is useful for action recognition. Just as we discussed above, model can recognize the actions by recognition some specific objects. But when action is confirmed, this RGB information is not so useful for adverb recognition and the high dimension noise will degrades the performance.

When using expression knowledge, $mAP$ values raise about 0.9 point which validates the expectation. And we analysis the results on the 51 adverbs in detail. We find that actually for the adverbs like ``heavily" and ``slowly" the result is almost the same (for ``heavily" $AP$ changes from 5.783 to 5.802 in task 2). While for the adverbs describing the mood and attitude like ``happily", the expression really helps (for ``happily" $AP$ changes from 6.632 to 7.924 in task 2). So we can tell expression recognition is an important part of adverb recognition.

\vspace{-0.01in}
Comparing task1 and task2, we find that the results for adverb recognition are almost the same, which means the action information is not related to adverb recognition. Therefore, the risk that the model is conducting action recognition while expected to conduct HAA recognition (for example, the model recognizes the adverb ``sweetly" due to seeing ``kiss") is beingless. This is also the reason why these wonderful models for action recognition do not work well for HAA recognition.

\begin{table}
	\begin{center}
		\footnotesize
		\begin{tabular}{|c|c|c|c|c|c|c|}
			\hline
			\multirow{2}{*} & \multicolumn{2}{|c|}{mAP} & \multicolumn{2}{|c|}{Hit@1} & \multicolumn{2}{|c|}{Hit@5}\\
			\cline{2-7}
			& Act & Adv &  Act & Adv &  Act & Adv \\
			\hline\hline
			T1-F1 & 6.968 & 6.413& 36.482 & 38.633 & 66.776 & 73.898 \\
			T1-F1-e & 7.732  & \textbf{7.347} & 40.763 & 43.276 & 70.235 &  74.235\\
			T1-F2 & \textbf{7.887} & 6.434 & 42.146 & 40.833 & 69.124 & \textbf{74.100} \\
			T1-F2-e & 7.886 & 7.153 & \textbf{45.315} & \textbf{44.356} & \textbf{70.451} & 73.892 \\
			\hline
			T2-F1 & - & 6.521 & - & 31.252 & - & 74.903 \\
			T2-F1-e & - & 7.362 & - & \textbf{36.351} & - & \textbf{75.362} \\
			T2-F2 & - & 7.251 & - & 10.587 & - & 16.560 \\
			T2-F2-e & - & \textbf{7.459} & - & 12.251 & - & 17.358 \\

			\hline
		\end{tabular}
	\end{center}
	\vspace{-0.05in}
	\caption{PBLSTM results. ``T1-F1" means task 1 with feature 1. ``-e" means using expression knowledge. Task 2 doesn't recognize actions so ``Act'' does not have values in task 2.}
	\label{tab:res_PBLSTM}
\end{table}	
\vspace{-0.01in}
The results of the two-stream model are shown in Tab.~\ref{tab:res_TS}. ``-S" means the spatial stream. ``-M" means the motion stream. ``-F" is the fusion of the two streams. The analysis discussed above is still suitable for this model.

The spatial stream utilizes the RGB information which can tell what is in the video. This information is not so useful for adverb recognition. Even for human, when a person sees an image with a walking human, it is also difficult to tell whether the walking man/woman is free or in a hurry. Compared with spatial stream the motion stream which uses optical flow information shows a much better performance. Obviously, with the speed and the direction of each movement, it is easy for a model to recognize the adverbs.
 
After fusion the two stream, the performance raises a bit which is not as much as expectation. It is because the spatial stream really achieves a bad performance. In the table, we show the best average fusing result with 20\% weight for spatial stream and 80\% weight for motion stream. We also try the max fusion strategy which is bad, just as we have expected.

Finally the hybrid model results are shown in Tab.~\ref{tab:res_HM}. The model integrating the information of pose, expression, RGB and optical flow achieve the best performance. But still it is not a satisfactory model and we need further enquiry.

\begin{table}
	\begin{center}
		\footnotesize
		\begin{tabular}{|c|c|c|c|c|c|c|}
			\hline
			\multirow{2}{*} & \multicolumn{2}{|c|}{mAP} & \multicolumn{2}{|c|}{Hit@1} & \multicolumn{2}{|c|}{Hit@5}\\
			\cline{2-7}
			& Act & Adv &  Act & Adv &  Act & Adv \\
			\hline\hline
			T1-S & 3.806 & 6.246 & 2.140 & 6.140 & 15.400  & 24.850 \\
			T1-M & 3.953 & 6.657 & 6.630 & 23.390 & \textbf{24.760} & 53.610\\
			T1-F & 4.126 & 6.792 & 5.870 & 23.190 & 24.120 & \textbf{55.140}\\
			T1-F-e & \textbf{5.623} & \textbf{7.064} & \textbf{7.160} & \textbf{24.650} & 24.150 & 54.320\\
			\hline
			T2-S & - & 6.272 & - & 2.780 & - & 14.960\\
			T2-M & - & 6.251 & - & 4.350 & - &\textbf{20.170} \\
			T2-F & - & 6.841 & - & 4.420 & - & 20.140\\
			T2-F-e & - & \textbf{7.624} & - & \textbf{4.560} & - & 20.160\\
			
			\hline
		\end{tabular}
	\end{center}
	\caption{Two-stream Model results. ``-S'' means spatial stream. ``-M'' means motion stream. ``-F'' means fusion streams. Task 2 doesn't recognize actions so ``Act'' does not have values in task 2.}
	\label{tab:res_TS}
	
\end{table}

\begin{table}
	\begin{center}
		\footnotesize
		\begin{tabular}{|c|c|c|c|c|c|c|}
			\hline
			\multirow{2}{*} & \multicolumn{2}{|c|}{mAP} & \multicolumn{2}{|c|}{Hit@1} & \multicolumn{2}{|c|}{Hit@5}\\
			\cline{2-7}
			& Act & Adv &  Act & Adv &  Act & Adv \\
			\hline\hline
			T1-H & 8.103 & 9.235 & 28.135 & 34.292 & 53.329  & 64.325 \\		
			\hline
			T2-H & - & 9.738 & - & 27.321 & - & 45.329\\

			\hline
		\end{tabular}
	\end{center}
	\caption{Hybrid models results. Task 2 doesn't recognize actions so ``Act'' does not have values in task 2.}
	\label{tab:res_HM}
\end{table}

\subsubsection{Influence of Bounding Boxes Correction}
All of the above models use the tracking bounding boxes with human correction which is impossible in real word applications. So we decide to analysis how much this correction will affect the performance.

We choose the best settings of the above models and use the detection and tracking results without human correction to redo the experiments. The result is shown in Tab.~\ref{tab:res_AC}. ``-a'' means getting the tracking boundingboxes automatically.

We can see that the human correction doesn't affect the results too much. One reason is that the human detection and tracking algorithms are good enough for the adverb recognition problem. Some tracking-lost problems would not affect the performance. Of course, one must not lose sight of the fact that all mentioned models do not work well.

\begin{table}
	\begin{center}
		\scriptsize
		\begin{tabular}{|c|c|c|c|c|c|c|}
			\hline
			\multirow{2}{*} & \multicolumn{2}{|c|}{mAP} & \multicolumn{2}{|c|}{Hit@1} & \multicolumn{2}{|c|}{Hit@5}\\
			\cline{2-7}
			& Act & Adv &  Act & Adv &  Act & Adv \\
			\hline\hline
			T1-F-e & 5.623 & 7.064 &7.160&24.650 & 24.150 & 54.320\\
			T1-F-e-a & 5.634 & 6.993& 7.210& 24.557 & 24.069 & 54.310\\
			T1-F1-e & 7.732  & 7.347 & 40.763 & 43.276 & 70.235 &  74.235\\
			T1-F1-e-a & 7.729  & 7.336 & 40.49& 44.282 & 68.124& 73.125\\
			\hline
			T2-F-e & - & 7.624 & - & 4.560 & - & 20.160\\
			T2-F-e-a & -  & 7.633 &- & 4.558 &- & 20.249\\
			T2-F1-e & - & 7.362 & - & 36.351 & - & 75.362 \\
			T2-F1-e-a & -  & 7.359 &- & 36.218 &- & 75.349\\
			
			\hline
		\end{tabular}
	\end{center}
	\caption{Results with and without human correction. ``-a'' means the model gets the tracking boundingboxes automatically. Task 2 doesn't recognize actions so ``Act'' does not have values in task 2.}
	\label{tab:res_AC}
\end{table}

\section{Conclusions}
We established the first benchmark for recognizing human action adverbs: ADHA. This task is beyond the pattern recognition problems like action recognition. In ADHA, we labeled the actions from a common and describable action set, the adverbs from a semantically complete adverb set, and the human boundingboxes for each person in each video. With ADHA, we benchmarked several outstanding action recognition models. From the result, we can tell that action and adverb recognition have little relativity and using those models achieved unsatisfactory results. Moreover, we propose a hybrid model using RGB, optical flow, pose and expression knowledge and showed that it achieve better results on HAA recognition problem.

{\small
\bibliographystyle{ieee}
\bibliography{egbib}

\begin{thebibliography}{10}\itemsep=-1pt

\bibitem{abu2016youtube}
S.~Abu-El-Haija, N.~Kothari, J.~Lee, P.~Natsev, G.~Toderici, B.~Varadarajan,
  and S.~Vijayanarasimhan.
\newblock Youtube-8m: A large-scale video classification benchmark.
\newblock {\em arXiv preprint}, 2016.

\bibitem{davies2008corpus}
M.~Davies.
\newblock {\em The corpus of contemporary American English}.
\newblock BYE, Brigham Young University, 2008.

\bibitem{deng2009imagenet}
J.~Deng, W.~Dong, R.~Socher, L.-J. Li, K.~Li, and L.~Fei-Fei.
\newblock Imagenet: A large-scale hierarchical image database.
\newblock In {\em IEEE Conference on Computer Vision and Pattern Recognition},
  pages 248--255, 2009.

\bibitem{dhall2016emotiw}
A.~Dhall, R.~Goecke, J.~Joshi, J.~Hoey, and T.~Gedeon.
\newblock Emotiw 2016: Video and group-level emotion recognition challenges.
\newblock In {\em ACM International Conference on Multimodal Interaction},
  pages 427--432, 2016.

\bibitem{donahue2015long}
J.~Donahue, L.~Anne~Hendricks, S.~Guadarrama, M.~Rohrbach, S.~Venugopalan,
  K.~Saenko, and T.~Darrell.
\newblock Long-term recurrent convolutional networks for visual recognition and
  description.
\newblock In {\em IEEE conference on computer vision and pattern recognition},
  pages 2625--2634, 2015.

\bibitem{douglas2007humaine}
E.~Douglas-Cowie, R.~Cowie, I.~Sneddon, C.~Cox, O.~Lowry, M.~Mcrorie, J.-C.
  Martin, L.~Devillers, S.~Abrilian, A.~Batliner, et~al.
\newblock The humaine database: addressing the collection and annotation of
  naturalistic and induced emotional data.
\newblock {\em Affective computing and intelligent interaction}, pages
  488--500, 2007.

\bibitem{fan2016video}
Y.~Fan, X.~Lu, D.~Li, and Y.~Liu.
\newblock Video-based emotion recognition using cnn-rnn and c3d hybrid
  networks.
\newblock In {\em ACM International Conference on Multimodal Interaction},
  pages 445--450, 2016.

\bibitem{fang2017rmpe}
H.-S. Fang, S.~Xie, Y.-W. Tai, and C.~Lu.
\newblock {RMPE}: Regional multi-person pose estimation.
\newblock In {\em ICCV}, 2017.

\bibitem{farneback2003two}
G.~Farneb{\"a}ck.
\newblock Two-frame motion estimation based on polynomial expansion.
\newblock {\em Image analysis}, pages 363--370, 2003.

\bibitem{girshick2015fast}
R.~Girshick.
\newblock Fast r-cnn.
\newblock In {\em IEEE international conference on computer vision}, pages
  1440--1448, 2015.

\bibitem{guadarrama2013youtube2text}
S.~Guadarrama, N.~Krishnamoorthy, G.~Malkarnenkar, S.~Venugopalan, R.~Mooney,
  T.~Darrell, and K.~Saenko.
\newblock Youtube2text: Recognizing and describing arbitrary activities using
  semantic hierarchies and zero-shot recognition.
\newblock In {\em IEEE international conference on computer vision}, pages
  2712--2719, 2013.

\bibitem{he2016deep}
K.~He, X.~Zhang, S.~Ren, and J.~Sun.
\newblock Deep residual learning for image recognition.
\newblock In {\em Proceedings of the IEEE conference on computer vision and
  pattern recognition}, pages 770--778, 2016.

\bibitem{ji20133d}
S.~Ji, W.~Xu, M.~Yang, and K.~Yu.
\newblock 3d convolutional neural networks for human action recognition.
\newblock {\em IEEE transactions on pattern analysis and machine intelligence},
  35(1):221--231, 2013.

\bibitem{karpathy2014large}
A.~Karpathy, G.~Toderici, S.~Shetty, T.~Leung, R.~Sukthankar, and L.~Fei-Fei.
\newblock Large-scale video classification with convolutional neural networks.
\newblock In {\em IEEE conference on Computer Vision and Pattern Recognition},
  pages 1725--1732, 2014.

\bibitem{Krishna_2017_ICCV}
R.~Krishna, K.~Hata, F.~Ren, L.~Fei-Fei, and J.~Carlos~Niebles.
\newblock Dense-captioning events in videos.
\newblock In {\em The IEEE International Conference on Computer Vision (ICCV)},
  Oct 2017.

\bibitem{Kristan_2015_ICCV_Workshops}
M.~Kristan, J.~Matas, A.~Leonardis, M.~Felsberg, L.~Cehovin, G.~Fernandez,
  T.~Vojir, G.~Hager, G.~Nebehay, and R.~Pflugfelder.
\newblock The visual object tracking vot2015 challenge results.
\newblock In {\em The IEEE International Conference on Computer Vision (ICCV)
  Workshops}, December 2015.

\bibitem{kuehne2011hmdb}
H.~Kuehne, H.~Jhuang, E.~Garrote, T.~Poggio, and T.~Serre.
\newblock Hmdb: a large video database for human motion recognition.
\newblock In {\em IEEE International Conference on Computer Vision (ICCV)},
  pages 2556--2563, 2011.

\bibitem{laptev2005space}
I.~Laptev.
\newblock On space-time interest points.
\newblock {\em International journal of computer vision}, 64(2-3):107--123,
  2005.

\bibitem{laptev2008learning}
I.~Laptev, M.~Marszalek, C.~Schmid, and B.~Rozenfeld.
\newblock Learning realistic human actions from movies.
\newblock In {\em IEEE Conference on Computer Vision and Pattern
  Recognition(CVPR)}, pages 1--8, 2008.

\bibitem{maji2011action}
S.~Maji, L.~Bourdev, and J.~Malik.
\newblock Action recognition from a distributed representation of pose and
  appearance.
\newblock In {\em IEEE Conference on Computer Vision and Pattern Recognition
  (CVPR)}, pages 3177--3184, 2011.

\bibitem{nam2015mdnet}
H.~Nam and B.~Han.
\newblock Learning multi-domain convolutional neural networks for visual
  tracking.
\newblock {\em CoRR}, abs/1510.07945, 2015.

\bibitem{newell2016stacked}
A.~Newell, K.~Yang, and J.~Deng.
\newblock Stacked hourglass networks for human pose estimation.
\newblock In {\em European Conference on Computer Vision}, pages 483--499,
  2016.

\bibitem{ng2015deep}
H.-W. Ng, V.~D. Nguyen, V.~Vonikakis, and S.~Winkler.
\newblock Deep learning for emotion recognition on small datasets using
  transfer learning.
\newblock In {\em ACM on international conference on multimodal interaction},
  pages 443--449, 2015.

\bibitem{pan2004gcap}
J.-Y. Pan, H.-J. Yang, C.~Faloutsos, and P.~Duygulu.
\newblock Gcap: Graph-based automatic image captioning.
\newblock In {\em Computer Vision and Pattern Recognition Workshop, 2004.
  CVPRW'04. Conference on}, pages 146--146, 2004.

\bibitem{petrosino2010toward}
A.~Petrosino and K.~Gold.
\newblock Toward fast mapping for robot adjective learning.
\newblock In {\em 2010 AAAI Fall Symposium Series}, 2010.

\bibitem{redmon2016yolo9000}
J.~Redmon and A.~Farhadi.
\newblock Yolo9000: better, faster, stronger.
\newblock {\em arXiv preprint}, 2016.

\bibitem{ren2015faster}
S.~Ren, K.~He, R.~Girshick, and J.~Sun.
\newblock Faster r-cnn: Towards real-time object detection with region proposal
  networks.
\newblock In {\em Advances in neural information processing systems}, pages
  91--99, 2015.

\bibitem{simonyan2014two}
K.~Simonyan and A.~Zisserman.
\newblock Two-stream convolutional networks for action recognition in videos.
\newblock In {\em Advances in neural information processing systems}, pages
  568--576, 2014.

\bibitem{soomro2012ucf101}
K.~Soomro, A.~R. Zamir, and M.~Shah.
\newblock Ucf101: A dataset of 101 human actions classes from videos in the
  wild.
\newblock {\em arXiv preprint}, 2012.

\bibitem{srivastava2015unsupervised}
N.~Srivastava, E.~Mansimov, and R.~Salakhudinov.
\newblock Unsupervised learning of video representations using lstms.
\newblock In {\em International Conference on Machine Learning}, pages
  843--852, 2015.

\bibitem{szegedy2016rethinking}
C.~Szegedy, V.~Vanhoucke, S.~Ioffe, J.~Shlens, and Z.~Wojna.
\newblock Rethinking the inception architecture for computer vision.
\newblock In {\em IEEE Conference on Computer Vision and Pattern Recognition},
  pages 2818--2826, 2016.

\bibitem{valstar2010induced}
M.~Valstar and M.~Pantic.
\newblock Induced disgust, happiness and surprise: an addition to the mmi
  facial expression database.
\newblock In {\em Workshop on EMOTION (satellite of LREC): Corpora for Research
  on Emotion and Affect}, page~65, 2010.

\bibitem{wang2011action}
H.~Wang, A.~Kl{\"a}ser, C.~Schmid, and C.-L. Liu.
\newblock Action recognition by dense trajectories.
\newblock In {\em IEEE Conference on Computer Vision and Pattern Recognition
  (CVPR)}, pages 3169--3176, 2011.

\bibitem{wang2016temporal}
L.~Wang, Y.~Xiong, Z.~Wang, Y.~Qiao, D.~Lin, X.~Tang, and L.~Van~Gool.
\newblock Temporal segment networks: Towards good practices for deep action
  recognition.
\newblock In {\em European Conference on Computer Vision}, pages 20--36, 2016.

\bibitem{white1991adverb}
L.~White.
\newblock Adverb placement in second language acquisition: Some effects of
  positive and negative evidence in the classroom.
\newblock {\em Interlanguage studies bulletin (Utrecht)}, 7(2):133--161, 1991.

\bibitem{wu2015modeling}
Z.~Wu, X.~Wang, Y.-G. Jiang, H.~Ye, and X.~Xue.
\newblock Modeling spatial-temporal clues in a hybrid deep learning framework
  for video classification.
\newblock In {\em ACM international conference on Multimedia}, pages 461--470,
  2015.

\bibitem{xu2016msr}
J.~Xu, T.~Mei, T.~Yao, and Y.~Rui.
\newblock Msr-vtt: A large video description dataset for bridging video and
  language.
\newblock In {\em IEEE Conference on Computer Vision and Pattern Recognition},
  pages 5288--5296, 2016.

\bibitem{xu2015show}
K.~Xu, J.~Ba, R.~Kiros, K.~Cho, A.~Courville, R.~Salakhudinov, R.~Zemel, and
  Y.~Bengio.
\newblock Show, attend and tell: Neural image caption generation with visual
  attention.
\newblock In {\em International Conference on Machine Learning}, pages
  2048--2057, 2015.

\bibitem{yue2015beyond}
J.~Yue-Hei~Ng, M.~Hausknecht, S.~Vijayanarasimhan, O.~Vinyals, R.~Monga, and
  G.~Toderici.
\newblock Beyond short snippets: Deep networks for video classification.
\newblock In {\em IEEE conference on computer vision and pattern recognition},
  pages 4694--4702, 2015.

\end{thebibliography}
}

\end{document}